\documentclass[letterpaper,twocolumn,10pt]{article}
\usepackage{usenix2019_v3}
\usepackage{hyperref}

\usepackage[normalem]{ulem}

\usepackage{breakurl}
\usepackage[normalem]{ulem}
\usepackage{endnotes}
\usepackage{url}

\usepackage{tabularx}
\usepackage{subcaption}

\usepackage{pifont}
\usepackage{caption}
\usepackage{epsfig,multirow,makecell,color}
\usepackage{amsmath}
\usepackage{filecontents}
\usepackage{amsfonts}
\usepackage{fdsymbol}
\usepackage[linesnumbered,boxed,lined,ruled,vlined]{algorithm2e}
\usepackage{listings}
\usepackage{xcolor}
\usepackage{pifont}
\usepackage{colortbl}
\usepackage{balance}
\usepackage{graphicx}
\usepackage{epstopdf}
\usepackage{float}
\usepackage[compact]{titlesec}
\usepackage{wrapfig}
\usepackage[export]{adjustbox}
\usepackage{datetime}
\usepackage{cleveref}
\usepackage{bbding}
\usepackage{fancyhdr}
\usepackage{enumitem}
\usepackage{thmbox}
\usepackage[labelfont=bf,font=normal,skip=1pt]{caption}
\usepackage{nccmath}

\usepackage{tikz}
\usepackage{listings}
\usepackage{lstautogobble}
\usepackage{booktabs}
\usepackage{multirow}
\usepackage{array}
\newcolumntype{P}[1]{>{\centering\arraybackslash}p{#1}}

\newcommand{\xmark}{\ding{55}}%
\newcommand{\sys}{\mbox{PowerInfer-2}}

\newcommand{\ignore}[1]{}

\begin{document}

\title{{\sys}: Fast Large Language Model Inference on a Smartphone}

\author{\rm Zhenliang Xue*,\; Yixin Song*,\; Zeyu Mi\Envelope,\; Xinrui Zheng,\; Yubin Xia,\; and Haibo Chen\\[5pt]
{\normalsize {Institute of Parallel and Distributed Systems (IPADS), Shanghai Jiao Tong University}}}

\maketitle

\renewcommand*{\thefootnote}{{\fnsymbol{footnote}}}
\footnotetext[1]{ Co-first authors.}
\renewcommand*{\thefootnote}{{\Envelope}}
\footnotetext[2]{ Corresponding author: Zeyu Mi (\burl{yzmizeyu@sjtu.edu.cn}).}
\renewcommand{\thefootnote}{\arabic{footnote}}
\setcounter{footnote}{0}

\begin{abstract}
Large language models (LLMs) on smartphones enable real-time AI assistance
and privacy-preserving, offline operation. However, resource constraints of smartphones
limit current deployments to small language models (SLMs), significantly compromising their capabilities.
This paper introduces \textit{\sys}, a smartphone-based framework that enables fast inference for LLMs exceeding the memory capacity.
The key insight is decomposing matrix operations into neuron clusters as the basic processing unit,
which enables flexible scheduling and efficient I/O-computation pipelining.
{\sys} leverages this neuron-cluster-based design in both computation and storage.
For computation, neuron clusters with dense activations are processed on NPU,
while sparse clusters use CPU.
The storage engine provides a fine-grained pipeline mechanism that coordinates cluster-level computation and I/O operations,
enhanced by a segmented neuron cache to reduce I/O activities.
{\sys} achieves up to a 27.8$\times$ speed increase compared to state-of-the-art frameworks.
{\sys} is the first system to serve a 47B LLM on a smartphone, achieving 11.68 tokens/s.
Notably, these performance improvements preserve model quality with negligible accuracy degradation.
\end{abstract}

\section{Introduction}
\label{sec:intro}

Large Language Models (LLMs) have revolutionized our daily lives and work patterns through their remarkable ability to process and generate human-like text.
While state-of-the-art LLMs such as GPT-4~\cite{ChatGPT} and Claude-3.5~\cite{Claude} operate on powerful data center GPUs (e.g., NVIDIA H100~\cite{H100} and GB200~\cite{GB200}),
there is an increasingly growing interest in deploying LLMs directly on ubiquitous smartphones.
This trend is evidenced by major tech companies' recent initiatives in mobile LLM integration:
Apple has launched Apple Intelligence~\cite{apple-intelligence},
Google has introduced Gemini Nano for Pixel devices~\cite{Gemini-nano},
and smartphone manufacturers including Samsung, Huawei, and OPPO have incorporated various LLM solutions into their smartphones~\cite{samsung-llm,huawei-llm,oppo-llm}.
The surge of interest in mobile LLM deployment stems from three key advantages:
it enables instantaneous AI assistance across system functions by eliminating network latency,
enhances user privacy through local data processing,
and ensures reliable performance regardless of network conditions.

However, mobile LLM deployment faces a critical challenge:
due to smartphones' limited computing and memory resources,
only small language models (SLMs) can be deployed,
significantly compromising model quality and capabilities.
For example, Google's Gemini Nano~\cite{Gemini-nano}, specifically designed for smartphones, contains only 3.25B parameters and
requires less than 2GB of memory, which is substantially smaller than models deployed on high-end PCs and servers (typically ranging from tens to hundreds of billions of parameters).
This model size constraint severely limits the model's capabilities,
as larger models consistently demonstrate superior performance
according to the ``scaling law''~\cite{kaplan2020scaling}.

To address these limitations, recent research focuses on system-level optimizations that leverage sparse activation patterns,
with solutions like PowerInfer~\cite{song2023powerinfer} and LLMFlash~\cite{alizadeh2024llm} demonstrating promising results in high-end PC environments.
However, these PC-centric solutions suffer from severe execution speed degradation on smartphones (i.e., approximately 70\% slower decoding speed for 7B models)
due to two key hardware gaps between smartphones and high-end PCs:
\vspace{-1ex}
\begin{itemize}
\item \textit{The Sparse Computing Gap:}
Smartphones' mobile NPUs, while offering superior performance for dense computations compared to mobile CPUs,
lack dedicated hardware support for sparse operations and struggle with unstructured sparse computations.
In contrast, high-end PCs' discrete GPUs efficiently handle both dense and sparse matrix operations compared to CPUs.
This hardware limitation significantly impacts the effectiveness of PowerInfer and LLMFlash's sparse computation approaches on smartphones,
as sparse computations are forced to run on CPUs.
Such CPU-only execution not only underutilizes the NPU's computational capabilities
but also fails to fully exploit the high memory bandwidth available in smartphones' unified memory architecture (UMA).
\vspace{-1ex}

\item \textit{The Storage Performance Gap:}
The Universal Flash Storage (UFS) in smartphones faces severe performance limitations compared to PC's NVMe SSDs,
achieving merely half the bandwidth for sequential reads (4GB/s vs. 7.5GB/s),
while for random reads, this performance gap widens dramatically to 12$\times$ (100K IOPS vs. 1,200K IOPS).
Consequently, despite its specialized I/O optimizations, LLMFlash still cannot overcome these fundamental storage bandwidth limitations,
resulting in frequent computational unit idling during I/O operations.
\end{itemize}
\vspace{-1ex}

This paper presents \textit{\sys}, an LLM-serving framework that efficiently runs large language models on smartphones,
achieving high inference speeds even for models exceeding available device memory.
Unlike existing solutions primarily targeting high-end PC environments,
\sys's core innovation lies in its fine-grained neuron cluster abstraction for mobile LLM inference.
A neuron cluster represents a collection of neurons exhibiting the same activation pattern,
and its neuron number is dynamically determined at runtime based on the target computing units.
This fine-grained abstraction enables both optimal utilization of heterogeneous computing resources
and efficient coordination between computation and I/O operations.

Based on the flexible neuron cluster abstraction,
{\sys} provides two key principles that redesign LLM inference to address smartphone-specific hardware constraints (e.g., NPU-CPU heterogeneity, UFS limitations).
First, given smartphones' limited computing capabilities compared to PCs,
it is crucial to maximize performance by intelligently distributing tasks between NPU and CPU based on their respective computational strengths.
Specifically, {\sys}'s \textit{Sparsity-Aware Adaptation} simultaneously assigns neuron clusters of different sizes to NPUs for efficient dense matrix operations and CPUs for flexible sparse computations.
This parallel execution not only maximizes computational throughput but also optimizes memory bandwidth utilization.
Moreover, {\sys} redistributes the ratio of neuron clusters between NPU and CPU to dynamically adapt to changes in sparsity patterns across different batch sizes,
which is a critical optimization overlooked by previous systems.

The second principle, \textit{I/O-Aware Orchestration}, addresses a fundamental challenge in mobile LLM inference:
the mismatch between traditional matrix-based computation granularity and mobile storage performance.
Conventional matrix-level operations often leave computing units idle while waiting for slow UFS I/O,
severely underutilizing available computing resources.
To solve this, {\sys} leverages its neuron cluster-based computation model to efficiently pipeline matrix operations,
which brings two key benefits:
First, it enables efficient overlap between computation and I/O, effectively hiding storage access latency.
Second, it allows the system to choose specialized I/O access patterns for different neuron clusters,
improving storage bandwidth utilization during inference.

We implemented {\sys} by extending PowerInfer~\cite{song2023powerinfer}, adding 12K lines of code.
{\sys} has been evaluated on two smartphones:
OnePlus 12 and Ace 2, all featuring heterogeneous Qualcomm XPUs with 24GB and 16GB DRAM.
Our system fundamentally breaks through the SLM constraints on smartphones,
enabling deployment of large language models that were previously impossible to run on mobile devices.
Specifically, {\sys} supports various LLM architectures and model sizes (from 7B to 47B).
Our evaluation shows three key improvements:
(1) Performance: 24.6$\times$ (up to 27.8$\times$) over llama.cpp and 3.84$\times$ average speedup (up to 4.63$\times$) over LLMFlash on OnePlus 12;
(2) Model size: First system to run large models like TurboSparse-Mixtral-47B on mobile, at 11.68 tokens/s;
(3) Memory: 40\% less memory usage for 7B models while matching llama.cpp and MLC-LLM speed.
Notably, these improvements maintain model quality with minimal accuracy impact.
\section{Background and Motivation}
\label{sec:bg}

\subsection{LLM Inference Background}
\label{subsec:bg_inference}
LLM inference operates in two distinct phases: prefill and decoding.
In the prefill phase, the model processes the entire user prompt in parallel,
producing the initial token.
Subsequently, in the decoding phase, the model generates text through an autoregressive process,
where each newly generated token becomes the input for producing the next one,
continuing until either the desired output length is reached or an end-of-sequence (EOS) token is encountered.

Contemporary LLMs predominantly utilize a decoder-only transformer architecture,
consisting of multiple transformer layers.
Each layer incorporates an attention mechanism and a Feed-Forward Network (FFN).
With the adoption of Group Query Attention~\cite{roumeliotis2023llama} in recent models,
the FFN component has become the dominant factor in model size,
accounting for approximately 80\% of total parameters in models like Llama3-8B, Qwen2-7B, and Mistral-7B.
This dominance makes FFN design choices crucial for inference performance.
Notably, ReLU-family activation functions~\cite{agarap2019deep, shazeer2020glu, zhang2024relu2}
have gained prominence in this context:
they not only achieve comparable model performance to non-ReLU alternatives,
but also introduce significant sparsity characteristics~\cite{zhang2022moefication, li2023lazy, luo2024sparsinglawlargelanguage},
as many neurons naturally become inactive due to their negligible contribution to the final output.

\subsection{LLM Inference with Dynamic Sparsity}

\begin{figure}[ht]
    \includegraphics[width=0.95\linewidth]{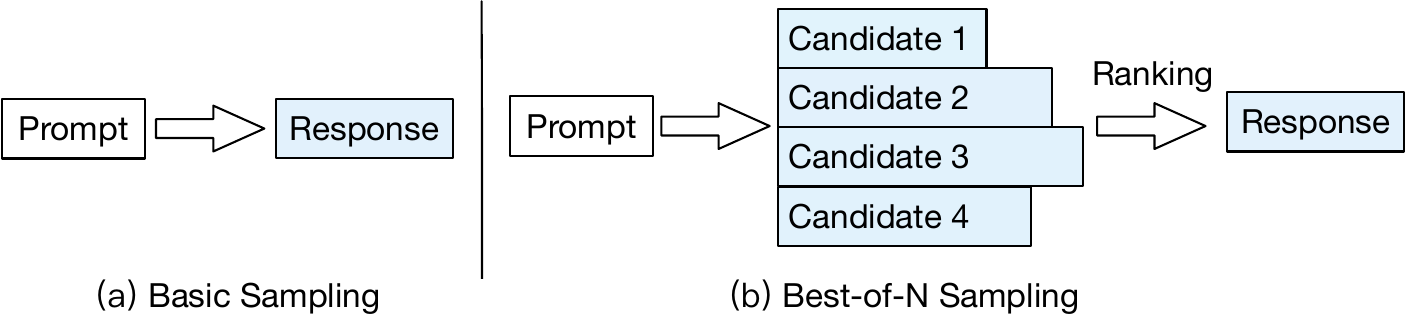}
    \caption{\small{Comparison of sampling strategies used in LLM inference.
    (a) Basic sampling generates a single response directly from the prompt.
    (b) Best-of-N sampling generates multiple candidate response sequences for one prompt,
    and selects the final response through a ranking process.}}
    \label{fig:sampling}
    \vspace{-3mm}
\end{figure}

Prior works~\cite{liu2023deja, zhang2024relu2, song2024prosparse, song2023powerinfer, alizadeh2024llm}
have demonstrated that FFN block activations can be predicted before computation.
PowerInfer~\cite{song2023powerinfer}, for instance, employs a \textit{static} approach,
distributing hot and cold neurons between GPU and CPU based on predicted activation patterns.
However, this approach overlooks an important aspect of modern LLM deployments:
the dynamic sparsity patterns introduced by advanced decoding strategies.
While traditional methods generate tokens sequentially using basic sampling (Fig.\ref{fig:sampling}-a), 
recent innovations have introduced advanced decoding methods that generate and evaluate multiple candidate sequences in parallel. 
As illustrated in Fig.\ref{fig:sampling}-b, techniques such as Best-of-N (BoN) sampling~\cite{lightman2023let, snell2024scaling} 
generate multiple candidates and select the best response. 
Similar parallel generation strategies are also employed in Monte Carlo Tree Search (MCTS)\cite{zhang2024accessing} and map-reduce paradigms\cite{zhou2024mapreduce}.
% GPT-o1-like sampling~\cite{o1}

\begin{figure}[ht]
    \centering
    \vspace{-0mm}
    \includegraphics[width=0.9\linewidth]{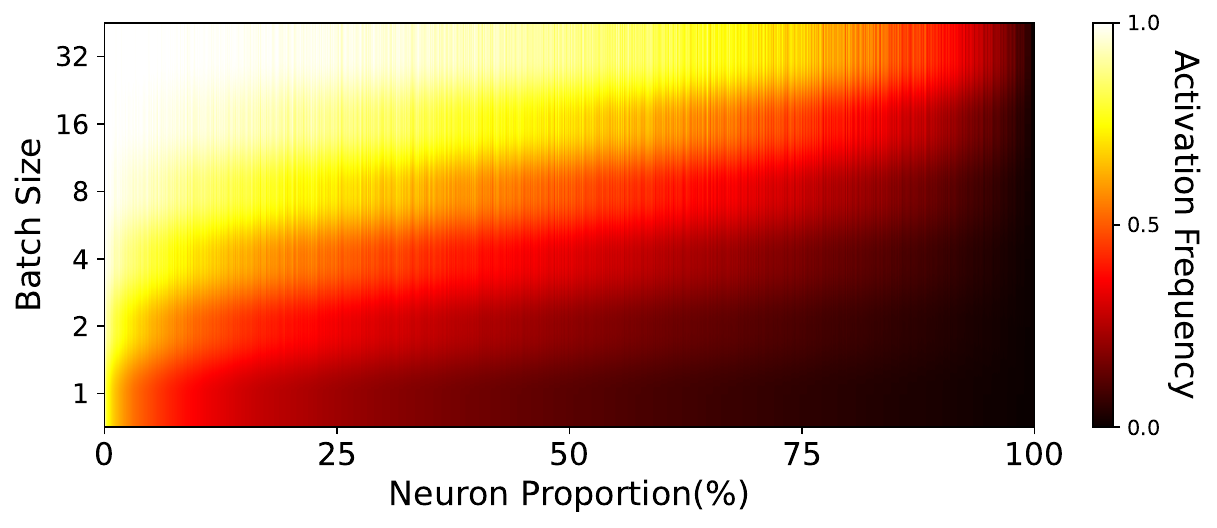}
    \caption{\small{
        Neuron activation patterns in layer 10 of Bamboo-7B under different batch sizes.
        The X-axis represents the proportion of neurons (sorted by activation frequency), 
        and the Y-axis shows different batch sizes. 
        Darker red indicates lower activation frequency. 
    }}
    \label{fig:batch-size-activation}
    \vspace{-5mm}
\end{figure}

As shown in Fig.\ref{fig:batch-size-activation},
parallel token generation leads to dynamic sparsity patterns in neuron activations\footnote{A neuron is considered activated when triggered by at least one token in the batch.} during inference.
Our analysis uncovers two key insights that current LLM frameworks fail to exploit:
First, while large batch sizes exhibit consistent ``hot spots'' of activated neurons,
single-batch processing demonstrates higher sparsity with more dispersed, irregular activation patterns.
Second, during decoding, the effective batch size exhibits dynamic fluctuations
due to varying sequence termination points.
The proportion of highly-activated neurons (depicted in white in Fig.\ref{fig:batch-size-activation})
demonstrates a dramatic correlation with batch size,
escalating from less than 1\% at batch size 1 to approximately 75\% at batch size 32.
These two insights not only indicate that an efficient inference system
should exploit sparsity patterns between hot and cold neurons,
but also needs to dynamically adapt to their shifting distributions.

\subsection{Mobile Hardware Characteristics}
\label{subsec:compute_characteristics}

Mobile devices present fundamentally different hardware characteristics from PCs,
manifesting in two key aspects that directly impact LLM inference:
(1) heterogeneous computing capabilities with distinct sparse computation characteristics, and
(2) distinct storage architecture with unique I/O characteristics.

\subsubsection{Computing Unit Characteristics}
Unlike high-end PC environments where discrete GPUs consistently dominate LLM computations,
smartphones feature multiple computing units (including CPU, GPU, NPU, etc.) with distinct performance characteristics,
particularly in handling sparse computations.
Taking OnePlus 12 (equipped with a Qualcomm Snapdragon 8 Gen 3 processor) as a case study,
we analyze how each processor type handles dense versus sparse computations in LLM inference tasks.

\begin{figure}[ht]
    % \resizebox{1\columnwidth}{!}{
    \subfloat[] {
        \includegraphics[width=0.48\linewidth]{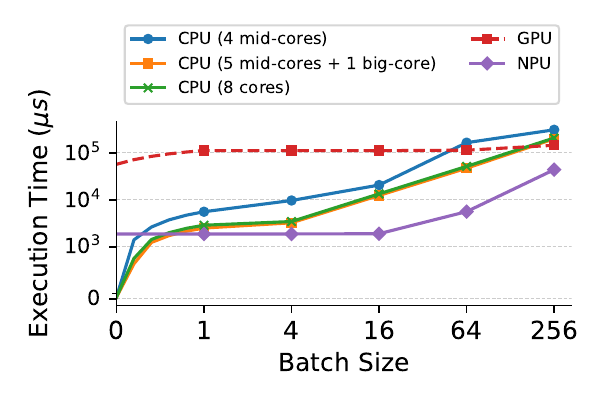}
    }
    % \hspace{-5mm}
    % \vspace{-5mm}
    \subfloat[] {
        \includegraphics[width=0.48\linewidth]{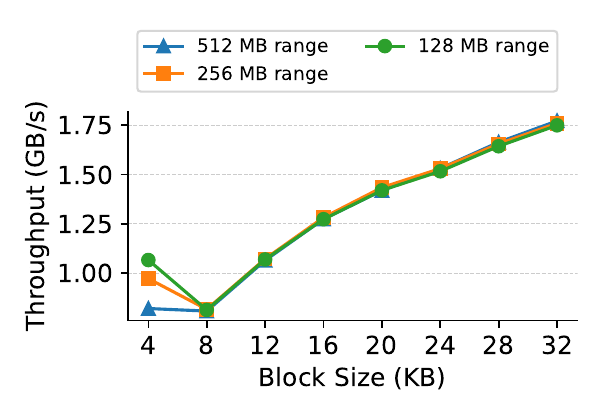}
    }
    % }
    % \caption{\textbf{Comparison between FlexGen and llama.cpp}
    % \vspace{-3mm}
    \caption{\small{(a) Comparison of execution times for matrix-vector multiplication operations with varying computational loads across CPU, GPU, and NPU platforms.
    The test involves multiplying a 14336×4096 matrix with vectors under different batch sizes.
    The X-axis represents the batch size while the Y-axis shows execution time.
    (b) Random read throughput performance for 4KB operations across different block sizes and data ranges.
    The X-axis depicts block size, while the Y-axis represents throughput.}}
    \label{fig:npu_cpu_matmul_test}
    \vspace{-2mm}
\end{figure}

\noindent\textbf{CPU's Advantage in Sparse Computing.}
Mobile CPUs utilize big.LITTLE technology~\cite{biglittle},
combining high-performance ``big'' cores with high-efficiency ``little'' cores.
The Snapdragon 8 Gen 3 features a 1+5+2 configuration: one big-core for peak performance,
five mid-cores for balanced performance, and two little-cores for energy efficiency.
Our experiments reveal that mobile CPUs, particularly the big and middle cores,
excel at handling sparse matrix-vector computations.
When handling sparse computations (matrices with batch size less than 1),
six CPU cores demonstrate faster performance compared to NPUs and GPUs, as illustrated in Fig.\ref{fig:npu_cpu_matmul_test}-a.

\noindent\textbf{NPU's Dense Computing Strength.}
The Neural Processing Unit (NPU), while efficient for dense computations,
shows limitations in sparse operations.
As shown in Fig.\ref{fig:npu_cpu_matmul_test}-a, NPUs excel at large batch processing,
performing significantly faster than CPUs and GPUs at high batch sizes.
Therefore, for a 7B INT4 model,
NPU achieves a throughput of 770 tokens/s for the prefill phase\footnote{Qualcomm's NPUs typically employ coarse-grained per-channel quantization that leads to accuracy degradation.
This issue can be fixed by adopting a mixed-precision quantization approach inspired by AWQ~\cite{lin2024awq, chen2024mixq},
which preserves outlier weights while applying INT4 quantization to the remaining components,
effectively maintaining model accuracy.},
significantly outperforming GPU's 37 tokens/s and CPU's 8 token/s.
However, our implementation of sparse computation primitives on Qualcomm's NPU
actually performs slower than CPU implementations, which is stark contrast to PC platforms
where GPUs excel at both dense and sparse computations.

\noindent\textbf{GPU's Limited Role.}
Mobile GPUs show consistent performance limitations in our tests.
Fig.\ref{fig:npu_cpu_matmul_test}-a demonstrates that for matrix-vector operations,
GPU consistently performs slower than both NPU and CPU,
with only about 50\% of GPU kernel time spent on actual computation.
Moreover, since GPUs are primarily responsible for rendering tasks,
utilizing them for inference would significantly impact the device's rendering frame rate~\cite{li2024transformerlitehighefficiencydeploymentlarge}.
Therefore, due to these performance limitations and rendering resource contention,
our private communications with mobile manufacturers
reveal that they avoid using GPUs for LLM inference in production systems.

\noindent\textbf{XPU and Memory Bandwidth Sharing.}
Smartphone's unified memory architecture (UMA) also
differs significantly from PC's separate VRAM and storage design.
Under this architecture, all processors share the same memory space,
enabling efficient memory bandwidth sharing.
Our measurements on OnePlus 12 running a 7B parameter model show that
memory bandwidth varies significantly based on processor utilization.
When performing matrix multiplication operations,
CPU-only computations achieve 43.9GB/s bandwidth,
while NPU-only computations reach 56GB/s.
However, when coordinating both processors simultaneously,
the aggregate memory bandwidth increases to 59.6GB/s.
This suggests that mobile LLM inference frameworks should leverage heterogeneous processors simultaneously
to maximize the utilization of shared memory bandwidth.

\subsubsection{Storage I/O Characteristics}
\label{subsec:storage_characteristics}
The second key challenge arises from the fundamental differences between mobile and PC storage architectures.
Mobile devices primarily use Universal Flash Storage (UFS), which exhibits markedly different I/O characteristics
compared to the NVMe SSDs found in PCs. Through our comprehensive analysis of UFS 4.0,
we identified four distinctive performance characteristics that significantly impact LLM inference:

\noindent\textbf{Block Size Impact.}
    Read bandwidth varies significantly with block size.
    For sequential reads, bandwidth ranges from 450 MB/s with 4KB blocks to 4 GB/s with 512KB blocks.
    Random read performance shows similar variation, reaching 3.5 GB/s with 512KB blocks
    but dropping significantly with smaller blocks.

\noindent\textbf{Data Range Sensitivity.}
    Random read performance is highly sensitive to the scope of the read range.
    As shown in Fig.\ref{fig:npu_cpu_matmul_test}-b, 4KB random reads within a 128MB range achieve 1 GB/s,
    while the same operations across a 512MB range drop below 850 MB/s.
    This characteristic is particularly pronounced for small block sizes,
    making data locality crucial for performance optimization.

\noindent\textbf{CPU Core Dependencies.}
    Storage performance is tightly coupled with the CPU core handling I/O operations.
    As shown in Table~\ref{tab:read-tput-different-cores}, using a big-core (3.3GHz) achieves 1 GB/s for 4KB random reads,
    while a little-core (2.2GHz) only reaches 760 MB/s.
    This correlation stems from the CPU's role in processing UFS driver operations,
    including interrupt handling and queue management.

 \noindent\textbf{Limited Concurrency.}
    Unlike NVMe SSDs with multiple command queues,
    UFS storage has a single command queue, limiting internal concurrency.
    Our tests show that using multiple cores for I/O operations can degrade performance by up to 40\%
    due to command queue contention.

\begin{table}[t]
    \caption{\small{4KB random read throughputs on 128MB data range with different core setups.}}
    \label{tab:read-tput-different-cores}
    % \vspace{-4mm}
    \resizebox{\linewidth}{!}{
        \scriptsize{
            \setlength{\arrayrulewidth}{0.01pt}  % 调整线的粗细，可以根据需要修改数值
            \fontsize{2pt}{2.2pt}\selectfont
                \begin{tabular}{c|c}
                    \hline
                    Core Setup & Throughput (MB/s) \\ \hline
                    big-core (3.3GHz) & 1,076.10 \\ \hline
                    mid-core (3GHz) & 1,007.95 \\ \hline
                    little-core (2.2GHz) & 761.87 \\ \hline
                \end{tabular}
        }
    }
    \vspace{-5mm}
\end{table}

\subsection{Limitations of Existing Solutions}
\label{subsec:existing_solutions}

These mobile hardware characteristics pose significant challenges for existing LLM inference frameworks.
To quantitatively analyze the performance bottlenecks of existing solutions on mobile platforms,
we implemented and evaluated both PowerInfer and LLMFlash on a OnePlus 12 smartphone running the Mistral 7B model.
We set a small memory budget for the evaluation and offloaded 50\% of FFN weights to flash storage.
Since PowerInfer was originally designed without flash storage support,
we extended its design to handle cases where model weights exceed available memory.
Both systems use Asynchronous I/O (AIO) to access cold neurons stored in flash storage.

\begin{table}[H]
    \caption{Decoding speeds and FFN performance in of Mistral 7B on existing methods w/wo offloading.}
    \label{tab:existing_solutions_performance}
    \centering
    \resizebox{0.48\textwidth}{!}{
        \begin{tabular}{l|cc|ccc}
        \toprule

        \multirow{2}{*}{Solutions} & \multicolumn{2}{c|}{In Memory} & \multicolumn{3}{c}{Offloading 50\% FFN Weight} \\ 
                    \cline{2-6}
                    & \makecell[c]{Decoding\\Speed}     & \makecell[c]{Memory\\Bandwidth} & \makecell[c]{Decoding\\Speed}  & \makecell[c]{I/O\\Overhead} & \makecell[c]{CPU\\Utilization} \\
                    \hline
        PowerInfer  & 12.4 tok/s                        & 41.1 GB/s                       & 1.4 tok/s            & 81.9\%                      & 31\%                        \\
        LLMFlash    & 12.9 tok/s                        & 43.9 GB/s                       & 2.3 tok/s            & 76.7\%                      & 41\%                        \\ 
        \bottomrule
        \end{tabular}
    }
    \vspace{-3mm}
\end{table}

Our experimental results demonstrate severe performance degradation when offloading 50\% of FFN weights to flash storage.
As shown in Table~\ref{tab:existing_solutions_performance}, PowerInfer experiences a 89\% decoding speed reduction, while LLMFlash suffers a 82\% decline.
This significant performance gap can be attributed to two fundamental challenges in the mobile computing environment.
First, existing solutions' computational strategies are heavily dependent on hardware-accelerated sparse operations.
While this approach is effective on PC platforms with powerful GPUs,
it becomes problematic on mobile devices where NPUs lack efficient sparse computation capabilities,
ultimately forcing all sparse computations onto the CPU.
Our measurements from Table~\ref{tab:existing_solutions_performance} reveal that even with weights stored in memory,
mobile memory bandwidth utilization reaches only approximately 43.9 GB/s due to CPU cores' limited computing capabilities,
which is far below 59.6GB/s when both NPU and CPU are fully engaged.
Second, the significantly lower I/O bandwidth of mobile UFS storage (compared to PC's NVMe SSDs)
creates a major performance bottleneck, with I/O overhead accounting for 82\% and 77\% of total latency
in PowerInfer and LLMFlash respectively.
The slow I/O operations force computing units to frequently idle while waiting for I/O operations,
with CPU utilization dropping to 41\% during weight loading phases.
\section{{\sys} Overview}
\label{sec:overview}

\subsection{Neuron Cluster and Design Principles}
This paper proposes neuron clusters as the basic unit for mobile LLM inference.
A neuron cluster consists of a group of neurons from the same FFN layer that exhibit similar activation patterns.
The cluster size is adaptively determined and categorized based on both hardware characteristics and activation frequencies:
larger hot clusters comprising frequently activated neurons are allocated for NPU-accelerated dense computations,
while smaller cold clusters containing rarely activated neurons are handled by CPU-driven sparse computations.

{\sys} builds upon this abstraction with two design principles.
First, \textbf{Sparsity-Aware Adaptation} maximizes hardware utilization through intelligent workload distribution.
During prefill phase, the system leverages NPU for efficient dense operations.
During decoding phase, it dynamically allocates workload between CPU and NPU based on sparsity patterns and batch sizes.
Second, \textbf{I/O-Aware Orchestration} optimizes UFS storage access through fine-grained pipelining
and differentiated I/O strategies for various types of LLM weights.

\subsection{Overall Architecture}

\begin{figure}[t]
    \begin{minipage}{1\linewidth}
        \centering\includegraphics[scale=1.1]{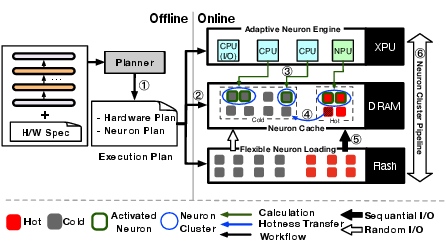}
    \end{minipage}
    \begin{minipage}{1\linewidth}
    \caption{\small{The architecture overview of {\sys}}.}
    \label{fig:arch}
    \end{minipage} % \\[-10pt]
    \vspace{-1cm}
\end{figure}

Fig.\ref{fig:arch} presents the overall architecture of {\sys},
which consists of an offline planning phase and an online inference phase.
Due to the dynamic sparsity patterns and heterogeneous computing characteristics of mobile platforms,
the offline phase needs to analyze both model characteristics and hardware capabilities to generate optimized execution plans.
During online inference, {\sys} proposes four techniques and mechanisms ($\S$\ref{sec:design}) to deliver efficient inference on mobile devices.

\noindent\textbf{Offline Planning:}
The planner first conducts offline analysis by running the model
to understand activation patterns across different batch sizes.
It classifies neurons into hot clusters and cold clusters
through a comprehensive process that considers both model characteristics and hardware capabilities.
Based on this analysis, it generates an execution plan (Step \ding{172}) including a neuron plan that guides how to split neurons into clusters,
and a hardware plan that guides how to configure the hardware resources for {\sys} (Step \ding{173}).

\noindent\textbf{Online Inference:}
During inference, an adaptive neuron engine (Step \ding{174}) efficiently orchestrates workload distribution 
between NPU and CPU based on the neuron plan and dynamic sparsity patterns.
The engine implements two specialized strategies: During the prefill phase, it uses an NPU-centric approach where the NPU 
handles dense matrix operations while a dedicated CPU core performs parallel weight loading.
For the decoding phase, the engine adopts a CPU-NPU hybrid strategy that 
optimally distributes workload - assigning hot neuron clusters to the NPU while delegating cold clusters to the CPU.
Similar to PowerInfer and LLMFlash, the CPU side features an online predictor to minimize computation overhead
by selectively calculating only those neurons predicted to be activated.
To accommodate dynamic sparsity patterns,
the engine continuously monitors and adjusts neuron cluster allocation between NPU and CPU as batch sizes evolve over time (Step \ding{175}).

Prior to initiating inference computations,
the computing engine retrieves neuron weights from a neuron cache
specifically designed to exploit distinct access patterns of different LLM weights.
For weights not present in the cache,
{\sys} introduces an adaptive neuron loading mechanism that uses differentiated I/O strategies to maximize efficiency (Step \ding{176}).
To mitigate the performance impact of random reads, {\sys} introduces a neuron cluster pipeline mechanism (Step \ding{177}).
For instance, upon completing the processing of a neuron cluster from matrix A,
CPU cores immediately transition to processing pre-cached neuron clusters from matrix B.
This operation executes concurrently with ongoing I/O operations for matrix A's in-flash neuron clusters,
enabling optimal parallelization between computation and I/O operations.
\section{Detailed Design of Online Inference}
\label{sec:design}

\subsection{Adaptive Neuron Engine}
\label{subsec:engine}

{\sys} introduces an adaptive neuron engine that proposes flexible computation strategies
based on different inference stages and dynamic sparsity, fully leveraging the computational capabilities of mobile XPUs.

\subsubsection{NPU-Centric Prefill}
In the prefill phase, all prompt tokens are processed concurrently.
Even though individual tokens exhibit high sparsity and activate distinct neurons,
the overall sparsity decreases significantly when aggregating these activations across the batch.
Consequently, {\sys} merges all neurons into a large cluster during the prefill stage
rather than using predictors to calculate activated neurons.
Given NPU's superior performance in large batch matrix operations compared to CPU,
{\sys} primarily leverages NPU for prefill computations.
Although CPU cores do not participate in matrix calculations during prefill,
the neuron engine leverage a CPU core to load all weights sequentially from flash into memory.

\begin{figure}[t]
    \begin{minipage}{1\linewidth}
        \centering\includegraphics[scale=1.3]{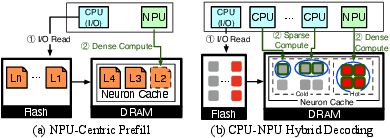}
    \end{minipage}
    \begin{minipage}{1\linewidth}
    \caption{\small{Two computing workflows for prefill and decoding phases.
    (a) The prefill phase uses an NPU-centric workflow that leverages NPU for computation;
    (b) The decoding phase employs a CPU-NPU hybrid workflow for FFN computation where NPU handles dense computations for hot neurons while CPU cores process sparse computations for cold neurons,
    with their processing ratio automatically adjusting to match the dynamic sparsity patterns caused by varying batch sizes.
    The attention computation is handled entirely by NPU but not shown in the figure.}}
    \label{fig:computing-workflow}
    \end{minipage} % \\[-10pt]
    \vspace{-3mm}
\end{figure}

Fig.\ref{fig:computing-workflow}-a shows how CPUs and NPU work together for prefill-phase inference. 
The NPU processes all LLM layers sequentially via its powerful dense matrix operations.
The NPU shares limited memory with the CPU, so the CPU preloads matrix weights into this shared memory before NPU computation. 
{\sys} uses a big-core to asynchronously preload weights for the next layer into the neuron cache. 
Since the prefill stage involves dense matrix calculations, sequential reads are used to load large data blocks, maximizing UFS's I/O bandwidth.

\subsubsection{Hybrid CPU-NPU Decoding}
\label{subsec:cpu_npu_decoding}
During decoding, {\sys} adopts distinct strategies for attention and FFN computations (Fig.\ref{fig:computing-workflow}-b).
For attention blocks, which are inherently dense but relatively small,
{\sys} partitions and distributes individual attention heads between CPU and NPU,
with each XPU processing its assigned heads in parallel.

For FFN blocks that dominate computation,
{\sys} adopts a hybrid approach where NPU handles dense hot neurons while CPU processes sparse cold neurons.
The neuron engine splits an FFN matrix along the neuron dimension (rows),
with the split ratio dynamically adjusting based on batch size.
For example,
when batch size is 1, it divides a typical FFN matrix (14336×4096)
into two equal sub-matrices (7168×4096) for NPU and CPU respectively.
As batch size increases, the NPU portion grows larger to handle more dense computations.
The NPU performs dense matrix multiplication on its portion,
while CPU cores process cold neurons with a predictor-based approach~\cite{song2023powerinfer}.
The CPU further divides its workload into smaller chunks (e.g., 512×4096) for parallel processing,
utilizing ARM Neon extensions for vectorized computation.
After both XPUs complete their computations,
the results are merged by the CPU side to form the final output.

\subsubsection{Dynamic CPU-NPU Adjustment}
To handle dynamic sparsity patterns efficiently, 
the adaptive neuron engine dynamically adjusts the CPU-NPU computation ratio based on runtime batch sizes.
CPU cores monitor batch size changes by tracking the completion and creation of decoding sequences.
For large batch sizes with multiple concurrent sequences,
sparsity is low due to high neuron activation overlap.
In this case, NPU handles most computations (about 70\%) while CPU processes the remaining sparse part (about 30\%).
As sequences complete over time and batch size decreases, sparsity increases,
shifting more workload to CPU cores (up to 50\% for batch size 1) while NPU focuses on remaining hot neurons.

Given NPU's static graph execution model, adjusting its computational load requires loading new computation graphs.
During the offline phase, {\sys} prepares multiple NPU computation graphs,
each optimized for a specific batch size and corresponding hot neuron ratio.
These graphs differ in their operator shapes based on the number of participating neurons.
When the engine detects the need for NPU load adjustment during runtime,
it initiates an asynchronous graph loading process while NPU is computing the attention block.
The new graph, typically only 10KB in size, is loaded into NPU memory
and becomes active immediately after attention computation completes.
This asynchronous loading mechanism completely overlaps with NPU computation,
ensuring seamless transitions between different computational configurations
without introducing additional overhead.

\subsection{In-Memory Neuron Cache}
\label{subsec:neuron_cache}

Efficient in-memory caching is crucial for optimizing inference performance by minimizing the frequency of I/O operations.
While prior work like LLMFlash~\cite{alizadeh2024llm} proposes bundling co-activated neurons to reduce I/O operations,
this approach faces two key limitations on mobile devices:
First, it overlooks the skewed activation patterns where a small set of hot neurons dominates the connections,
leading to redundant loading of these frequently activated neurons across different bundles.
Second, our analysis shows that after excluding hot neurons,
the co-activation probability among remaining neurons drops below 20\%,
making the bundling strategy ineffective for I/O reduction.

To overcome these limitations, {\sys} implements a temperature-based caching strategy with two objectives:
eliminating redundant loading by explicitly distinguishing hot and cold neurons, and
optimizing cache efficiency for cold neurons through individual management rather than bundling, given their low co-activation probability.
Specifically, the cache consists of three regions:
(1) a fixed region for attention weights and KV cache that are preloaded and retained;
(2) a hot region for NPU dense computations; and
(3) a cold region for CPU sparse computations.

The hot and cold regions employ different organizational strategies.
The hot region groups neurons into dense matrices, combining Gate, Up, and Down sub-matrices mentioned in $\S$\ref{subsec:cpu_npu_decoding} into a unified hot neuron cluster.
In contrast, the cold region manages neurons individually.
Both regions use LRU-based eviction, but at different granularities:
cluster-level for the hot region and neuron-level for the cold region.
Evicted weights from both regions are simply discarded from memory without being written back to storage.

When batch sizes change, {\sys} dynamically adjusts the hot-cold region ratio.
For larger batches, the hot region expands to support increased NPU computations,
while the cold region shrinks by evicting the least active neurons via LRU policy.
Conversely, for smaller batches, the hot region also discards the least active neuron clusters to make room for the cold region.

\subsection{Neuron-Cluster-Level Pipeline}
\label{subsec:neuron_pipeline}

\begin{figure}[t]
    \begin{minipage}{1\linewidth}
        \centering\includegraphics[scale=1.2]{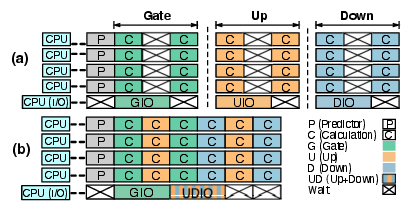}
    \end{minipage}
    \begin{minipage}{1\linewidth}
    \caption{\small{Two types of pipelines that combine matrix-vector multiplications and I/O operations on five cores (4 calculation cores and 1 I/O cores).
    Assuming that there are 8 neuron cluster per matrix, 4 of them are in memory and the other 4 are in flash before the computation starts. 
    (a) The matrix-level pipeline separates the pipeline into isolated matrix units; (b) The neuron-cluster-level pipeline in {\sys} breaks the matrix barrier and mixs their calculation and I/O operations in the neuron cluster granularity.}}
    \label{fig:neuron-pipeline}
    \end{minipage}
    \vspace{-5mm}
\end{figure}

Equipped with the neuron cache, I/O operations for uncached weights remain inevitable.
While attention blocks and hot neurons benefit from efficient sequential reads,
cold neurons require random I/O operations due to their sparse activation patterns.
This results in significantly reduced bandwidth due to UFS's poor random read performance,
creating a major bottleneck that impacts systems like PowerInfer and LLMFlash on mobile devices ($\S$\ref{subsec:existing_solutions}).

Therefore, {\sys} should hide I/O latency by overlapping computation with I/O operations.
A straightforward approach is matrix-level overlapping: processing cached neurons while simultaneously loading uncached neurons from storage.
However, this approach still has limitations. The system must wait for all neurons within a matrix to be available
before proceeding to the next matrix, especially when some neurons require lengthy I/O operations.
For example, as shown in Fig.\ref{fig:neuron-pipeline}-a, with a matrix containing 8 neuron clusters
(4 in memory, 4 in storage), CPU cores may still experience idle periods waiting for I/O completion,
despite partial overlapping of computation and I/O.

To minimize I/O latency, {\sys} introduces a neuron-cluster-level pipeline that tightly integrates computation and I/O activities,
following the \textbf{I/O-Aware Orchestration} principle.
Concretely, {\sys} breaks down the barriers between matrix computations;
as soon as one neuron cluster finishes computing, it immediately starts the computation of a neuron cluster in the next matrix that are in memory.
This mechanism overlaps the I/O operations within neuron cluster computations from multiple matrices,
effectively reducing waiting bubbles, as illustrated in Fig.\ref{fig:neuron-pipeline}-b.

{\sys} divides the execution process of a neuron cluster into 5 sequential stages,
which are: determining whether the rows/columns of the Gate, Up, and Down matrices is activated through the predictor (Pred),
reading the weights of the rows of the Gate matrix from storage (GIO),
calculating the product of the rows of the Gate matrix and the input vector (GC),
reading the rows/columns of the Up and Down matrices from storage (UDIO),
and calculating the product of the rows/columns of the Up and Down matrices with the input vector respectively (UDC).
{\sys} creates multiple computing threads and one I/O thread to handle the computations and I/O operations for these 5 stages, respectively.
The specific number of these threads and the cores on which they will execute are determined by the offline planner.

\subsection{Flexible Neuron Loading}
\label{subsec:neuron_load}

{\sys}'s I/O-aware orchestration employs distinct I/O strategies for different weight types.
For attention weights that are used densely during prefill,
{\sys} loads them entirely into memory along with other parameters using sequential reads,
and retains them in the attention region of the cache throughout inference.
For FFN hot neurons, {\sys} also utilizes sequential reads but employs asynchronous preloading
during attention computation to ensure immediate availability for FFN operations.
For cold neurons that exhibit sparse activation patterns,
{\sys} adopts an on-demand loading approach with small random reads.

To further optimize cold neuron loading efficiency,
{\sys} bundles correlated neurons across Gate, Up, and Down matrices.
While neurons within a single FFN matrix rarely co-activate after removing hot neurons,
corresponding neurons at the $i$-th position across these matrices show high correlation,
with an 80\% co-activation probability.
Thus, {\sys} organizes storage by neuron position rather than matrix structure,
storing weights of corresponding neurons from all three matrices as a single unit.

{\sys} further introduces distinct I/O loading strategies for different models,
considering their quantization methods and the inherent characteristics of UFS I/O.
For unquantized models like Mistral-7B-FP16, where each neuron occupies 8KB,
{\sys} performs single large random reads of 24KB to load the complete Gate-Up-Down bundle,
maximizing I/O bandwidth efficiency.
For 4-bit quantized models, each neuron bundle (Gate-Up-Down) requires 7.5KB
(2KB int4 weights + 0.5KB FP16 scales per matrix).
{\sys} aligns the bundle size to 8KB for storage efficiency, but strategically divides loading into two separate 4KB operations,
as empirical measurements demonstrate superior bandwidth utilization compared to a single 8KB random read ($\S$\ref{subsec:storage_characteristics}).

Moreover, while these bundles show 80\% co-activation rate, 
there is still a 20\% chance that bundled neurons do not activate together. 
Loading the entire bundle at once could waste bandwidth. 
For 4-bit quantized models, {\sys} adopts a two-phase loading strategy where it first loads the Gate matrix weights (4KB) after predictor confirms activation, 
and only loads the Up/Down weights (4KB) if the Gate neuron output is non-zero. 
This approach minimizes unnecessary I/O operations while maintaining performance.
\section{Execution Plan Generation}
\label{sec:planner}

To enable efficient LLM inference across diverse mobile platforms, {\sys} employs a targeted offline planning phase
that generates device-specific execution plans. This planning phase optimizes for different runtime scenarios
and dynamic batch sizes, automatically adapting neuron assignments between NPU and CPU based on each device's
unique hardware capabilities.

\textbf{Neuron Classification:}
The planner's primary task is to analyze and classify neurons based on their activation patterns
across different batch sizes and hardware I/O characteristics. This classification process involves:
(1) running the model on a diverse dataset (10M+ tokens from Wikipedia and RefinedWeb),
(2) tracking activation frequencies for each neuron under various batch sizes,
(3) analyzing hardware-specific I/O patterns and computation speeds, and
(4) categorizing neurons into hot clusters and cold clusters.

The classification considers two key factors:
First, the inherent activation patterns of neurons under different batch sizes.
Second, the hardware I/O capabilities of the target device.
Since hot neurons are asynchronously prefetched during the previous attention block's computation,
{\sys} carefully balances the number of hot neurons based on the available I/O bandwidth
and attention block computation time. This ensures that asynchronous I/O operations
are completely hidden within the attention computation process.

For example, during the prefill phase, dense activation patterns dominate,
leading to NPU processing assignments. For the decoding phase, the planner performs
detailed analysis across batch sizes (1-4). For larger batches (e.g., 3-4),
more neurons are classified as hot clusters due to denser activation patterns.
For smaller batches (1-2), more neurons are classified as cold clusters
due to sparse activation patterns. These assignments are further refined
based on the device's specific I/O characteristics to optimize the overlap
between computation and data movement.

\textbf{Batch-Adaptive Planning:}
Based on the neuron classification results, the planner prepares a series of execution strategies
optimized for different batch sizes. These strategies include
pre-generated static NPU computation graphs for common batch sizes, and I/O patterns tailored to each execution mode.
This pre-planning enables the runtime system to quickly adapt its execution strategy
when batch size changes occur, avoiding the overhead of real-time analysis
while maintaining optimal performance across different inference phases.

\textbf{Hardware-Aware Optimization:}
The planner also considers hardware capabilities when generating these batch-adaptive strategies.
It profiles the target device's NPU performance characteristics for different matrix sizes,
CPU capabilities for sparse computation,
memory bandwidth under various access patterns, and
storage I/O performance for different access sizes.
These hardware insights enable {\sys} to automatically adapt its execution strategy
to different mobile platforms, ensuring optimal performance across diverse devices.
\section{Implementation}
\label{sec:impl}

{\sys} is developed on top of PowerInfer~\cite{song2023powerinfer}, 
a state-of-the-art serving framework designed for sparsely-activated LLMs, 
by integrating an additional 12K lines of C++ code into PowerInfer~\cite{song2023powerinfer}. 
Since {\sys} depends on privileged system APIs (e.g., \texttt{mlock} that locks pages in memory) that need the root permission,
we built it on the Android~\cite{Android} platform.
Even though there is no need to alter the system kernel, 
a rooted Android system still provides us with considerable flexibility in developing and debugging our system. 
Furthermore, {\sys} is inherently designed with no modifications to the kernel,
making it easily portable to other operating systems, including iOS~\cite{IOS} platform.
\section{Evaluation}
\label{sec:eval}

\subsection{Experimental Setup}

\noindent\textbf{Hardware.} 
We evaluate on two OnePlus~\cite{oneplus} smartphones representing different performance tiers:
the high-end OnePlus 12 with flag-ship Snapdragon 8 Gen 3 SoC and the mid-end OnePlus Ace 2 with Snapdragon 8+ Gen 1 SoC.
Details are listed in Table~\ref{tab:devices}. 
\begin{table}[]
    \caption{
        Hardware specifications of smartphones we used in the evaluation.
        ``DRAM'' is the physical memory size.
        ``Available'' is the maximum memory size that can be occupied by an application.}
    \label{tab:devices}
\footnotesize

\scriptsize{
\begin{tabular}{l|l|l|l}
\hline
\textbf{Device Name} & \textbf{\begin{tabular}[c]{@{}l@{}}DRAM / Available \end{tabular}} & \textbf{Storage} & \textbf{SoC}  \\ \hline
OnePlus 12 & 24GB / 19GB & UFS 4.0     &   Snapdragon 8 Gen 3 \\ \hline
OnePlus Ace 2 & 16GB / 11GB  & UFS 3.1 &   Snapdragon 8+ Gen 1 \\ \hline
\end{tabular}
}
\end{table}

\noindent\textbf{Models:} 
We choose five LLMs of varying architectures and sizes,
namely Mistral(SiLU)-7B, sparse Qwen2-7B~\cite{yang2024qwen2}, Bamboo-7B~\cite{song2024turbo}, sparse Llama-13B~\cite{song2024prosparse},
and TurboSparse-Mixtral-47B~\cite{song2024turbo} (``Mixtral-47B'' in figures for short).

\noindent\textbf{Baselines:}
We compare {\sys} with four state-of-the-art LLM inference frameworks:
llama.cpp~\cite{llama.cpp}, QNN~\cite{QNN}, MLC-LLM~\cite{mlc-llm}, and LLMFlash~\cite{alizadeh2024llm}.
These frameworks represent different approaches to on-device inference:
llama.cpp is the most popular CPU-based framework that supports offloading weights to flash storage (via \texttt{mmap}),
and serves as the backend for many other frameworks like Ollama~\cite{ollama};
QNN is Qualcomm's commercial inference engine with sophisticated and proprietary NPU optimizations;
and MLC-LLM leverages mobile GPUs for acceleration.
LLMFlash is designed for high-end PC context and not open-sourced.
Therefore we implemented LLMFlash's key optimizations in llama.cpp based on the paper's description, including sparsity prediction, row-column bundling, and neuron data caching.

\noindent\textbf{Workloads:}
We evaluate using representative real-world LLM tasks including multi-turn dialogue~\cite{tunstall2023zephyr}, 
code generation~\cite{chen2021evaluating}, math problem solving~\cite{cobbe2021gsm8k}, 
and role play~\cite{wang2023rolellm}. These tasks are top representatives of real-world LLM tasks on the HuggingFace platform. 
Our evaluation covers prefill performance with 128 and 512 token prompts, decoding performance with up to 64 token prompts 
and 1,024 token outputs, as well as Best-of-N (BoN) generation speed. All experiments are averaged over 10 runs to 
account for variance.
\begin{figure}[!ht]
    \vspace{-3mm}
    \includegraphics[width=1\linewidth]{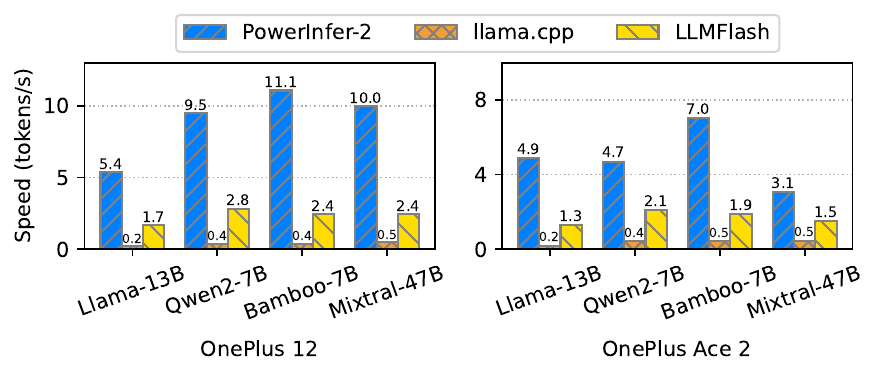}
    \caption{\small{
        Decoding speeds of {\sys}, llama.cpp and LLMFlash.
        The Y axis is the generation speed (tokens/s).
        50\% model weights of FFN blocks are offloaded to flash storage for all models
        except TurboSparse-Mixtral-47B on OnePlus Ace 2, which requires offloading at least 75\% of FFN weights.
    }}
    \label{fig:offload-decode}
    \vspace{-3mm}
\end{figure}
\subsection{Offloading-Based Performance}

In this section, we evaluate {\sys}'s decoding and prefill performance across different models under memory-constrained conditions.

\subsubsection{Decoding Performance}

We limit FFN weight placement in DRAM to 50\% across all models, except for TurboSparse-Mixtral-47B 
on OnePlus Ace 2 which requires 75\% offloading due to memory constraints. 
As shown in Fig.\ref{fig:offload-decode}, {\sys} demonstrates significant speedups across devices: 
on OnePlus 12, it achieves average speedups of 3.84$\times$ (up to 4.63$\times$) over LLMFlash 
and 24.6$\times$ (up to 27.8$\times$) over llama.cpp. 
Similarly, on OnePlus Ace 2, {\sys} delivers average speedups of 2.93$\times$ and 14.1$\times$ 
compared to LLMFlash and llama.cpp respectively.

\begin{table}[h]
\centering
\caption{Time proportion comparison between Compute and I/O for Bamboo-7B.}
\small  % 使用更小的字体
\setlength{\tabcolsep}{2.5pt}  % 减小列间距
\begin{tabular}{lcc}
\toprule
System & Compute & I/O \\
\midrule
{\sys} & 86.3\% & 13.7\% \\
LLMFlash & 23.3\% & 76.7\% \\
\bottomrule
\end{tabular}
\label{tab:io-critical}
\end{table}

When model size exceeds available DRAM, weight parameters must be dynamically swapped between flash storage and memory,
introducing substantial I/O overhead. Although LLMFlash's neuron cache mechanism achieves a 5.35$\times$ speedup
compared to llama.cpp's direct \texttt{mmap} approach, approximately 10\% of activated neurons still require
loading from flash storage, forcing computation to stall while waiting for random I/O completion.

{\sys} effectively addresses these challenges through two key innovations:
(1) an efficient neuron-cluster-level pipeline that overlaps I/O operations with computation, and 
(2) flexible bundle loading that maximizes I/O throughput.
These techniques effectively eliminate the I/O bottleneck, delivering superior performance across all evaluated models.
As demonstrated in Table~\ref{tab:io-critical}, our analysis of the critical path breakdown reveals
that while LLMFlash suffers from significant I/O overhead (76.7\%) due to frequent flash storage access
with only 23.3\% spent on actual computation,
{\sys} dramatically reduces I/O overhead to just 13.7\%, substantially improving overall system efficiency.

The acceleration ratio of {\sys} varies across models due to differences in activated parameters.
For instance, while TurboSparse-Mixtral-47B has 47B parameters in total,
its mixture-of-expert architecture and high sparsity result in only about 3B activated parameters per token,
similar to Bamboo-7B. This explains their comparable performance.
At 9.96 tokens/s, TurboSparse-Mixtral-47B still has room for improvement
through enlarged neuron cache, as discussed in $\S$\ref{sec:decode-speed-with-varying-memory}.
In contrast, Llama-13B's lower sparsity leads to nearly 2$\times$ more activated parameters than Bamboo-7B,
resulting in 2$\times$ slower performance.

\subsubsection{Prefill Performance}

\begin{figure}[]
    \vspace{-3mm}
    \includegraphics[width=1\linewidth]{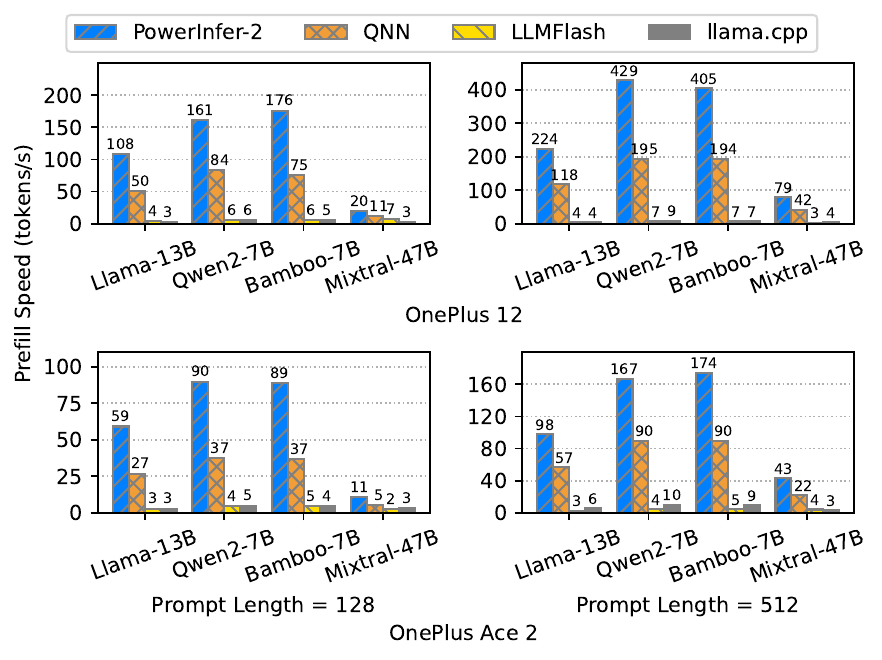}
    \caption{\small{
        Prefill speeds of {\sys}, QNN, llama.cpp and LLMFlash in offloading scenarios at prompt lengths of 128 and 512 tokens. 
        The X axis is the model. The Y axis is the prompt processing speed (tokens/s). 
        The placement of weights in FFN blocks is the same as Fig.\ref{fig:offload-decode}.
    }}
    \label{fig:offloading-prefill}
    \vspace{-3mm}
\end{figure}

In this section, we evaluate {\sys}'s prefill performance with 128 and 512 token prompts. 
Fig.\ref{fig:offloading-prefill} shows the prompt processing speeds across different systems.
With 512-token prompts, {\sys} achieves significant speedups:
48.97$\times$ over LLMFlash, 44.23$\times$ over llama.cpp, and 1.99$\times$ over QNN on OnePlus 12,
while showing 29.98$\times$ and 16.67$\times$ improvements on OnePlus Ace 2.
The results are similar for 128-token prompts, which reach up to 22.47$\times$ speedup over LLMFlash.

\begin{figure}[ht]
    \centering
    \vspace{-1mm}
    \includegraphics[width=0.8\linewidth]{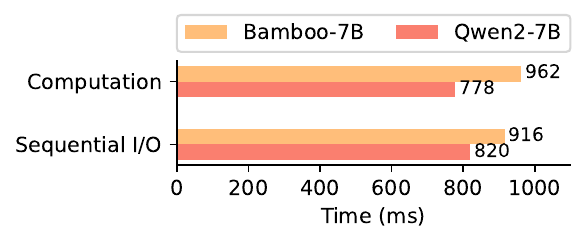}
    \caption{\small{
        Computation and sequential I/O time in a layer with prompt lengths of 512 tokens 
        at the prefill stage of Bamboo-7B and Qwen2-7B on OnePlus 12.
        The X axis is time used in milliseconds.
    }}
    \label{fig:prefill-bottleneck}
    \vspace{-2mm}
\end{figure}

These performance gains stem from {\sys}'s NPU-centric prefill and flexible neuron loading mechanism that preloads transformer layers using sequential I/O.
The NPU demonstrates superior performance for large batch sizes compared to CPU and GPU.
Additionally, {\sys} efficiently utilizes sequential I/O during prefill.
As batch sizes increase, neuron activation probability approaches 99.99\% in TurboSparse-Mixtral-47B with 128 batch size.
Therefore, {\sys} sequentially loads entire transformer layers using sequential I/O, which is 3$\times$ faster than random I/O.
Furthermore, weight loading is efficiently overlapped with computation through prefetching,
where weights for the next layer are loaded while computing the current layer.
As illustrated in Fig.\ref{fig:prefill-bottleneck}, I/O operations are completely overlapped with computation time through our pipelining strategy,
effectively hiding I/O latency.

\subsubsection{Performance with Various Memory Capacities}
\label{sec:decode-speed-with-varying-memory}

\begin{figure}[ht]
    \vspace{-1mm}
    \includegraphics[width=1\linewidth]{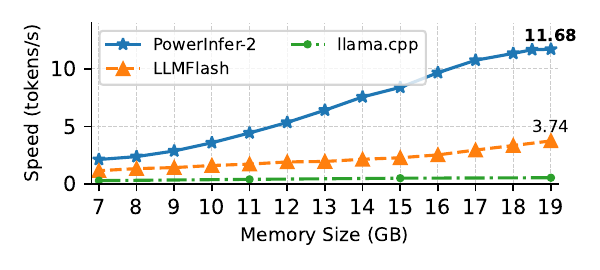}
    \caption{\small{
        Decoding speeds on various memory configurations with TurboSparse-Mixtral-47B on OnePlus 12.
    }}
    \label{fig:memory-speed}
    \vspace{-1mm}
\end{figure}

In real-world scenarios, smartphones run multiple applications simultaneously~\cite{mps}, 
leading to varying available memory for LLM inference.
We evaluate {\sys}'s adaptability across memory 
capacities from 7GB to 19GB. Fig.\ref{fig:memory-speed} shows the decoding speeds 
under different memory configurations for TurboSparse-Mixtral-47B.

With only 7GB available memory, {\sys} achieves 2.13 tokens/s. The memory is allocated 
as follows: 1GB for non-FFN layer weights\footnote{Including token embeddings, 
self-attention layers, KV cache, and the final language model head}, 2.6GB for predictor weights, 
2.7GB for FFN weights' quantization scales, and 300MB for runtime memory, totaling 
6.6GB. The remaining 400MB is used for neuron cache, which only stores 1.8\% of FFN 
weights. This small cache size means nearly all neurons must be fetched from flash 
storage for each token. Despite this limitation, {\sys} still outperforms LLMFlash 
by 1.84$\times$ through efficient neuron loading and neuron-cluster pipeline.
{\sys}'s decoding speed scales linearly with memory size up to 19GB, as I/O operations 
reduce proportionally with increased cache size. At maximum memory utilization (19GB), 
{\sys} reaches 11.68 tokens/s, which is 3.12$\times$ faster than LLMFlash and 21.2$\times$ 
faster than llama.cpp.

\subsubsection{Decoding Consistency Analysis}

\begin{figure}[]
    \vspace{-3mm}
    \includegraphics[width=1\linewidth]{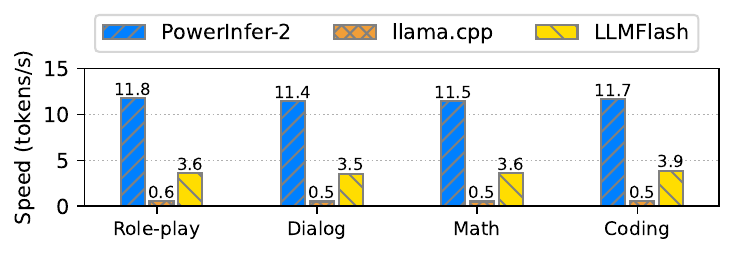}
    \caption{\small{
        Decoding performance on different downstream tasks of TurboSparse-Mixtral-47B on OnePlus 12.
        All available memory is used during decoding.
    }}
    \label{fig:tasks}
    \vspace{-3mm}
\end{figure}

To evaluate {\sys}'s performance consistency, 
we analyze decoding speeds across different tasks and examine token-level latency distribution.
For task-level analysis, 
we measure decoding speeds across four representative tasks: 
role-play, multi-turn dialogue, math problem solving, and code generation. 
As shown in Fig.\ref{fig:tasks}, 
{\sys} maintains consistent performance with at least 11.4 tokens/s across all tasks, 
demonstrating robust handling of diverse workloads. 
Minor speed variations occur due to task-dependent differences in model activation sparsity.

To analyze token-level latency distribution,
we measured the average, 50th percentile (P50), 90th percentile (P90), 
and 99th percentile (P99) decoding speeds across 1,024 tokens
for both TurboSparse-Mixtral-47B and Bamboo-7B models with 50\% of their FFN weights in DRAM.
TurboSparse-Mixtral-47B shows notable performance variations:
the slowest 10\% of tokens are generated 16.5\% below average speed, while P99 latency is 40.9\% slower than the mean.
These variations arise from differing activation patterns between consecutive tokens.
When tokens share activation patterns, they benefit from cached neurons,
avoiding costly flash storage access.
Our analysis reveals that while TurboSparse-Mixtral-47B maintains a low 3.5\% average cache miss rate,
the P99 miss rate reaches 18.9\%.
This significant variance in cache efficiency directly impacts token generation latency through increased neuron swapping between cache and flash storage.

\begin{table}[!t]
    \centering
    \caption{Decoding latencies of {\sys} in milliseconds when offloading 50\% of FFN weights.}
    \label{tab:inter-token}
    \resizebox{0.75\linewidth}{!}{
    \begin{tabular}{c|c|c}
    \toprule
    ~ & TurboSparse-Mixtral-47B  & Bamboo-7B \\
    \midrule
    Mean & 99.76 & 90.32 \\
    P50 & 97.42  & 86.88 \\
    P90 & 116.16  & 115.02 \\
    P99 & 140.56  & 162.02 \\
    \bottomrule
    \end{tabular}
    }
\end{table}

\subsubsection{SiLU-based LLM Performance}
\label{subsec:swiglu-performance}

\begin{table}[!t]
    \centering
    \caption{Generation speed (tokens/s) comparison for SiLU and ReLU-based LLMs when Offloading 50\% FFN weights.}
    \label{tab:performance_silu}
    \resizebox{0.8\linewidth}{!}{
    \scriptsize{
        \setlength{\arrayrulewidth}{0.1pt}
    \begin{tabular}{l|cccc}
    \toprule
    \textbf{Model} & \textbf{{\sys}} & \textbf{LLMFlash} & \textbf{Speedup} \\ \midrule
    Mistral(SiLU)-7B & 5.3 & 2.18 & 2.4$\times$ \\
    Bamboo-7B & 11.1 & 2.4 & 4.6$\times$ \\ \bottomrule
    \end{tabular}
    }
    }
    \vspace{-5mm}
\end{table}

While {\sys} achieves its remarkable acceleration on ReLU-based models due to their naturally high activation sparsity, 
it can also effectively accelerate SiLU-based architectures,
since recent studies (CATS~\cite{lee2024cats} and CHESS~\cite{he2024chess}) have shown that SiLU-based LLMs also exhibit approximately 50\% activation sparsity. 
As shown in Table~\ref{tab:performance_silu}, {\sys} achieves a 2.4$\times$ speedup on SiLU-based LLMs.
Though more modest than the gains seen with ReLU-based models, 
this improvement validates the versatility of {\sys}'s sparse computation mechanisms across different activation functions.
The relatively lower speedup can be attributed to the fundamental characteristics of SiLU-based architectures,
where a higher proportion of neurons remain active during inference,
creating a performance bottleneck in neuron loading operations.
\subsection{In-Memory Performance}

\begin{figure}[ht]
    \vspace{-1mm}
    \includegraphics[width=1\linewidth]{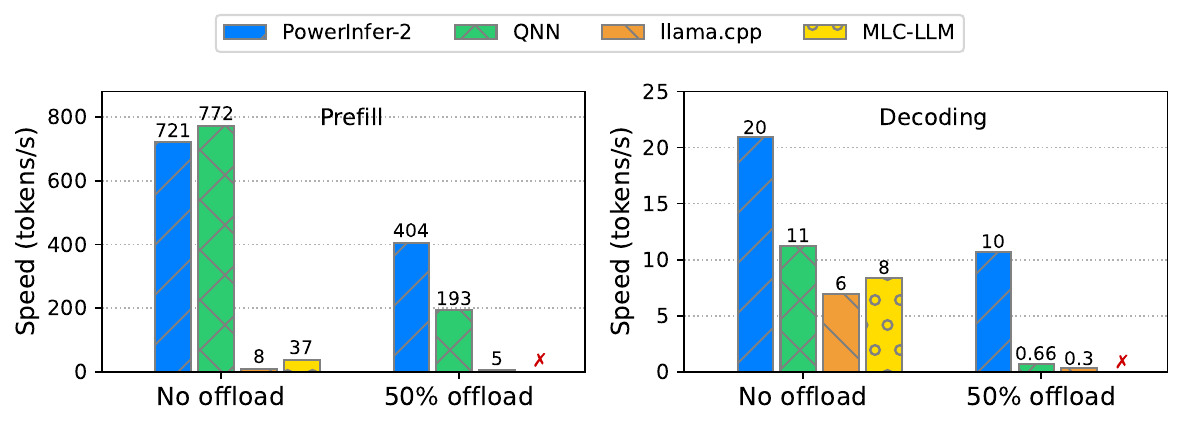}
    \caption{\small{
        Decoding speeds of {\sys}, llama.cpp, and MLC-LLM on Bamboo-7B with different offloading setups.
        ``50\% offload'' means 50\% model weights of FFN blocks are offloaded to flash storage.
        ``No offload'' means all model parameters are resident in memory.
        A red label of ``\xmark'' indicates an execution failure due to the lack of weight offloading support.
    }}
    \label{fig:in-memory-decode}
    \vspace{-1mm}
\end{figure}

This section evaluates Bamboo-7B's performance under sufficient memory conditions,
comparing against llama.cpp (CPU), MLC-LLM (GPU) and QNN (NPU) on smartphones,
as shown in Fig.\ref{fig:in-memory-decode}.

For the prefill phase, {\sys} fully leverages NPU acceleration, achieving comparable performance to QNN 
with both systems exceeding 700 tokens/s, which is a 32$\times$ speedup over other frameworks.
{\sys}'s NPU-centric approach in prefill stage ensures maximum hardware utilization,
though it is slightly slower than QNN due to mixed precision quantization used by {\sys},
a deliberate trade-off for accuracy preservation (will be discussed in $\S$\ref{subsec:accuracy-evaluation}).
With 50\% FFN weight offloading, {\sys} maintains 404.6 tokens/s throughput,
outperforming QNN through efficient pipelining that overlaps weight loading operations.
The performance gap compared to the in-memory setup is primarily due to the additional I/O overhead from flash storage access.

In the decoding phase, {\sys} demonstrates 2.24$\times$, 2.48$\times$ and 1.86$\times$ speedups over 
llama.cpp, MLC-LLM and QNN respectively. This superior performance comes from:
1) concurrent CPU-NPU computation improving memory bandwidth utilization from 43.9GB/s to 59.6GB/s, and
2) leveraging the model's activation sparsity to reduce FFN computations.

When offloading 50\% of FFN weights, {\sys} reduces memory usage by 1.5GB (40\%)
while maintaining performance comparable to llama.cpp and MLC-LLM.
This demonstrates {\sys}'s ability to optimize memory efficiency without compromising user experience,
even for models that already fit in memory.

\subsection{Best-of-N Sampling}

\begin{figure}[ht]
    \vspace{-1mm}
    \includegraphics[width=1\linewidth]{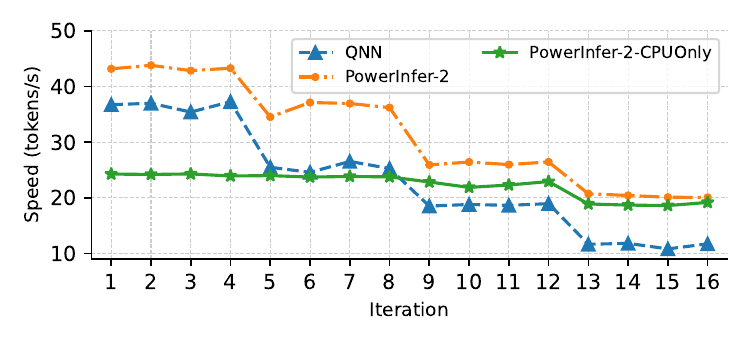}
    \caption{\small{
        Decoding speed curves of {\sys}, QNN, and {\sys}-CPUOnly on Bamboo-7B
        under the setting of Best-of-N (N=4). The batch size decreases by one every four iterations.
        X axis refers to iteration. Each iteration represents the generation of one token across N candidates in parallel.
        Y axis refers to the generation speed (tokens/s).
        All model parameters are resident in memory.
    }}
    \label{fig:BoN}
    \vspace{-1mm}
\end{figure}

Best-of-N (BoN) sampling, where N different response sequences are generated for one prompt and the best one is selected, 
is widely utilized in LLM applications. 
We evaluated Best-of-4 sampling on Bamboo-7B using three configurations: {\sys}, QNN, and 
{\sys}-CPUOnly ({\sys} with CPU-only decoding).
As shown in Fig.\ref{fig:BoN},
in the initial phase with all four samples generating simultaneously, {\sys} achieves 1.84$\times$ 
and 1.28$\times$ speedups over QNN and {\sys}-CPUOnly respectively. As samples complete and N 
decreases, {\sys} maintains a 1.42$\times$ performance advantage through dynamic CPU-NPU task 
dispatching based on sparsity patterns. When N reduces to 1, increased sparsity causes QNN's 
performance to drop below {\sys}-CPUOnly. However, {\sys} still achieves 1.1$\times$ and 
1.77$\times$ speedups over {\sys}-CPUOnly and QNN respectively, benefiting from its hybrid 
CPU/NPU computation strategy.

\subsection{Performance Breakdown}
\begin{figure}[ht]
    \centering
    \vspace{-1mm}
    \includegraphics[width=0.8\linewidth]{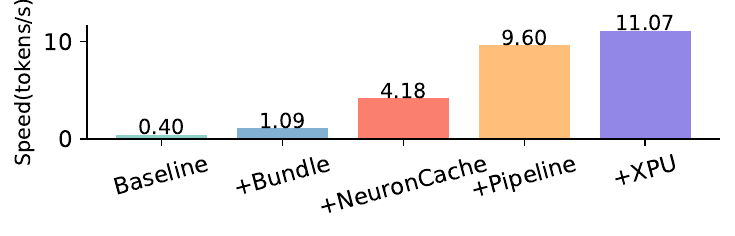}
    \caption{\small{Performance breakdown for decoding phase of the Bamboo-7B model on OnePlus 12, with 50\% FFN weights offloaded to flash storage.    }}
    \label{fig:ablation}
    \vspace{-1mm}
\end{figure}

To quantify each optimization's contribution,
we conducted an ablation study by incrementally enabling optimizations.
Starting from a baseline configuration, we progressively added Bundle, Neuron Cache, Neuron-Cluster-Level Pipeline, and XPU optimizations.
Fig.\ref{fig:ablation} shows the performance impact of each addition.

The baseline system utilizes CPU without any optimizations
and achieves 0.4 tokens/s without optimizations.
Adding Bundle optimization improves this to 1.1 tokens/s
by increasing I/O efficiency through bundled memory access operations.
Neuron Cache further brings a 2.82$\times$ speedup (4.181 tokens/s)
by achieving a 95\% cache hit rate,
significantly reducing Flash I/O bandwidth pressure.
Pipeline optimization adds another
1.29$\times$ improvement (9.60 tokens/s) through overlapped computation and memory access.
Finally, XPU optimization provides
an additional 15\% gain (11.07 tokens/s) by utilizing both CPU and XPU for concurrent memory access.

\subsection{Accuracy Evaluation}
\label{subsec:accuracy-evaluation}

We evaluate inference accuracy using OpenCompass~\cite{2023opencompass}, a widely-adopted benchmark suite,
focusing on INT4 quantization which is the de facto standard in real-world mobile deployments~\cite{lu2024bluelm, xiao2024large}.
As shown in Table~\ref{tb:acc}, {\sys} maintains comparable accuracy to llama.cpp across different models and tasks,
while significantly outperforming QNN.
The accuracy gap between frameworks stems from their different quantization approaches.
QNN uses per-channel quantization (one scale per weight matrix row), which poorly handles weights with outlier values.
llama.cpp achieves better accuracy by quantizing weights in groups of 32.
Since NPUs don't support group-wise quantization,
{\sys} adopts a hybrid quantization approach that handles outlier weights in INT8 precision
while applying INT4 per-channel quantization to remaining weights,
effectively preserving accuracy while maintaining fast inference speed similar to QNN.

\begin{table}[!t]
\centering
\caption{Comparison of LLM accuracy between different quantization methods on Bamboo 7B and Qwen2 7B models. 
The evaluation metrics include Arc Challenge~\cite{clark2018think}, Arc Easy, MMLU~\cite{hendrycks2020measuring}, and GSM8K~\cite{cobbe2021training} benchmarks.}
\label{tb:acc}
\resizebox{0.99\linewidth}{!}{
\begin{tabular}{l|c|ccccc}
\toprule
Model & Framework & Average & Arc-Challenge & Arc-Easy & MMLU & GSM8K \\
\midrule
\multirow{3}{*}{Qwen2 7B} 
& llama.cpp & 79.25 & 82.37 & 87.46 & 69.72 & 77.43 \\
& QNN & 56.93 & 65.76 & 80.25 & 59.45 & 22.26 \\
& {\sys} & 78.38 & 81.36 & 88.18 & 68.73 & 75.24 \\
\midrule
\multirow{3}{*}{Bamboo 7B}
& llama.cpp & 70.12 & 73.90 & 84.83 & 61.24 & 60.50 \\
& QNN & 63.26 & 69.15 & 78.31 & 57.00 & 48.59 \\
& {\sys} & 68.35 & 71.86 & 82.01 & 61.48 & 58.05 \\
\bottomrule
\end{tabular}
}
\vspace{-5mm}
\end{table}

\subsection{Energy Consumption}
\begin{table}[h]
\vspace{-3mm}
\centering
\caption{Energy consumption comparison between different frameworks.}
\small
\setlength{\tabcolsep}{2.5pt}
\begin{tabular}{lccc}
\hline
Framework & {\sys} & QNN & llama.cpp \\
\hline
Peak Power (W) &5.095 & 5.133  & 4.065 \\
Energy (J/token) & 0.257 & 0.373  & 0.672 \\
\hline
\end{tabular}
\label{tab:energy-comparison}
\vspace{-2mm}
\end{table}

We evaluate the energy efficiency of {\sys} through evaluation of end-to-end energy consumption during inference. Our primary metric is Joules per token (J/token), which quantifies the average energy required to generate each output token. To ensure representative results, we randomly sampled 100 prompts from the lmsys/lmsys-chat-1m~\cite{zheng2023lmsyschat1m} dataset, which contains real-world user interactions collected from chatbot deployments. The mean energy consumption results are shown in Table~\ref{tab:energy-comparison}.
{\sys} demonstrates superior energy efficiency, achieving 0.257 J/token, which represents a 31.1\% and 61.8\% reduction in energy consumption compared to QNN and llama.cpp, respectively. 
This enhanced efficiency is attributed to {\sys}'s ability to maintain similar peak power while achieving faster computation speeds, resulting in lower overall energy consumption.
\section{Related Work}
\label{sec:related}

\noindent\textbf{Resource-Efficient DNN/LLM Inference.}
The deployment of DNNs and LLMs on resource-constrained devices has gained significant traction recently~\cite{xu2024survey}.
While BAND~\cite{jeong2022band} demonstrates the feasibility of running multiple compact DNNs on a single XPU,
its methodology does not address the memory-flash hierarchy and
cannot scale to accommodate LLMs due to their substantial computational demands.
MLC-LLM~\cite{mlc-llm} represents a significant advancement in enabling GPU-accelerated LLM deployment on mobile devices.
Nevertheless, it is constrained by the requirement that models must fit entirely in memory.
Although mllm-NPU~\cite{xu2024empowering} leverages on-device NPU,
it lacks CPU-NPU co-processing and requires models to fit in memory,
limiting its use for larger models.
While STI~\cite{guo2023sti} introduces pipeline mechanisms for I/O-compute overlapping,
its capability is limited to BERT-scale models (0.33B parameters) on embedded devices,
making it insufficient for modern LLMs that typically exceed 1B parameters.
Various model compression techniques have emerged, including network pruning~\cite{ma2023llm,jaiswal2024emergence}, 
knowledge distillation~\cite{jung2023impossible}, and quantization~\cite{lin2024awq, chee2024quip},
all aimed at reducing model memory footprints.
These approaches complement {\sys} and can be integrated with it to further enhance the efficiency of LLM deployment on mobile devices.

\noindent\textbf{Speculative Decoding.}
Speculative decoding accelerates inference by leveraging a smaller model to predict tokens in parallel~\cite{li2024eagle, cai2024medusa, fu2024break}.
The technique employs a smaller model (e.g., 1B) to generate candidate tokens,
which are then validated in batch by a larger model (e.g., 13B).
SpecInfer~\cite{miao2023specinfer} demonstrates that this approach effectively reduces decoding iterations.
While speculative decoding and sparse activation are orthogonal optimization strategies,
their integration in memory-constrained XPU environments remains an open research challenge.

\section{Conclusion}
\label{sec:concl}

This paper presents {\sys}, a mobile-centric LLM inference framework that effectively runs LLMs on smartphones with limited memory.
Evaluation demonstrates that {\sys} achieves up to 27.8$\times$ speedup over existing solutions while supporting models as large as 47B with minimal accuracy loss.

\small{
\bibliographystyle{plain}
\bibliography{ms}
}

\balance
    
\end{document}